\documentclass[10pt,twocolumn,twoside] {IEEEtran}

\usepackage[section]{algorithm}
\usepackage{algorithmic}
\usepackage{array}
\usepackage{amsmath}
\usepackage{amssymb}
\usepackage{amsfonts}
\usepackage{amsthm}
\usepackage{booktabs}
\usepackage{color}
\usepackage{graphicx}
\usepackage{subfigure}
\usepackage{multirow}
\usepackage{threeparttable}
\usepackage{hyperref}
\usepackage{makecell}
\usepackage{fmtcount}

\newcolumntype{L}[1]{>{\raggedright\let\newline\\\arraybackslash\hspace{0pt}}m{#1}}
\newcolumntype{C}[1]{>{\centering\let\newline\\\arraybackslash\hspace{0pt}}m{#1}}
\newcolumntype{R}[1]{>{\raggedleft\let\newline\\\arraybackslash\hspace{0pt}}m{#1}}

\numberwithin{equation}{section}

\theoremstyle{remark}

\newcommand{\eg}{e.g.}

\newcommand{\R}{\mathbb{R}}

\newcommand{\scal}[2]{\left\langle #1,#2 \right\rangle}

\newcommand{\suml}[2]{\sum\nolimits_{#1}^{#2}}

\title{A higher-order MRF based variational model for multiplicative noise reduction} 
\author{Yunjin Chen,
  Wensen Feng,  Ren{\'e} Ranftl, Hong Qiao and Thomas Pock \thanks{Y.J. Chen, R. Ranftl and T. Pock are with the
    Institute for Computer Graphics and Vision, Graz University of
    Technology, Inffeldgasse 16, A-8010 Graz, Austria. 
e-mail: (\{cheny, ranftl, pock\}@icg.tugraz.at)}
  \thanks{Wensen Feng is with University of Science and Technology
Beijing, Beijing, 100190, China, also with the Institute of Automation, Chinese Academy of
Sciences, Beijing, 100190, China. e-mail: (sanmumuren@gmail.com)}
\thanks{Hong Qiao is with the Institute of Automation, Chinese Academy of
Sciences, Beijing, 100190, China. 
e-mail: (hong.qiao@mail.ia.ac.cn)}
\thanks{This work was supported by the
   Austrian Science Fund (FWF) under the China Scholarship Council
 (CSC) Scholarship Program and the START project BIVISION,
 No. Y729.}}
\begin{document}

\markboth{IEEE SIGNAL PROCESSING LETTERS,~Vol.~xx, No.~xx, 20xx}
{Chen \MakeLowercase{\textit{et al.}}: A higher-order MRF based variational model for multiplicative noise reduction}
\maketitle

\begin{abstract}
The Fields of Experts (FoE) image prior model, a filter-based higher-order Markov Random Fields (MRF) model, 
has been shown to be effective for many image restoration problems. 
Motivated by the successes of FoE-based approaches, in this letter we propose a novel variational model 
for multiplicative noise reduction based on the FoE image prior model. The resulting model corresponds to a non-convex 
minimization problem, which can be efficiently solved by a recently published non-convex optimization algorithm. 
Experimental results based on synthetic speckle noise and real synthetic aperture radar (SAR) images suggest that 
the performance of our proposed method is on par with the best published despeckling algorithm. 
Besides, our proposed model comes along with an additional advantage, that the inference is extremely efficient. 
{Our GPU based implementation takes less than 1s to produce state-of-the-art despeckling performance. }
\end{abstract}


%\IEEEpeerreviewmaketitle
%\vspace*{-0.25cm}
\section{Introduction}
Images generated by coherent imaging modalities, \eg, synthetic aperture radar (SAR), ultrasound and laser imaging, 
inevitably come with multiplicative noise (also known as speckle), due to the coherent nature of the scattering phenomena. 
The presence of this noise prevents us from interpreting valuable information of images, such as textures, edges and point 
target, and therefore speckle reduction is often a necessary preprocessing step for successful use of classical image processing 
algorithms involving image segmentation and classification. 
The topic of speckle noise reduction (despeckling) has attracted a lot of research attention since early 1980s 
\cite{lee1980digital, kuan1987adaptive}. 
At present it has been extensively studied. Roughly speaking, the major despeckling techniques fall into four categories: 
filtering based methods in (1) spatial domain; or (2) a transform domain, \eg, wavelet domain; (3) nonlocal filtering and 
(4) variational methods. 

Early filtering techniques in the spatial domain are developed under the minimum mean square error (MMSE) criterion 
\cite{lee1980digital}, and then progress to more sophisticated and promising maximum a 
posterior (MAP) approaches \cite{kuan1987adaptive}. Recently, bilateral filtering has also been 
modified for despeckling \cite{LiWHY13}. The emergence of wavelet transform in the early of 1990s, 
opened the way to a new generation of despeckling techniques. There were intensive studies of wavelet based 
despeckling approaches, see, for instance \cite{BianchiAA08, ArgentiBLA12} and references therein. 

The nonlocal approaches, which can take the advantage of self-similarity commonly present in natural as well as SAR images, 
have been already introduced to SAR despeckling \cite{DeledalleDT09, ParrilliPAV12, CozzolinoPSPV14}. 
By taking into account the peculiar features of multiplicative noise, the so-called SAR-BM3D algorithm \cite{ParrilliPAV12}, which is 
a SAR-oriented version of the well-known BM3D algorithm \cite{BM3D}, exhibits the best published despeckling performance at present. 

The last class of methods are variational ones, which minimize some appropriate energy functionals, consisting of 
a regularizer (also called image prior) and a data fitting term. Up to now, the well-known total variation (TV) has been 
widely used as a regularizer \cite{SteidlT10, YunW12, AubertA08}, and the total generalized variation (TGV) 
regularizer \cite{BrediesKP10} also has been investigated in a recent work \cite{wensen14}. 

There is a long history of research on regularization technique for image processing problems. 
The recently proposed Fields of Experts (FoE) \cite{RothFOE2009} 
image prior model (a higher-order filter-based MRF model), which explicitly characterizes the 
statistics properties of natural images, defines more effective variational models than hand-crafted regularizers, such as 
TV and TGV models. The variational model based on a discriminatively trained FoE prior 
can generate clear state-of-the-art performance for many image restoration problems, 
such as the additive white Gaussian noise (AWGN) denoising task, see for instance \cite{ChenPRB13, ChenRP14}. 
%In this work, study this issue with the obvious 
%goal of replicating the success it exhibits in the AWGN denoising context.

Motivated by the results of \cite{ChenPRB13, ChenRP14}, it is interesting to investigate the FoE prior based model for despeckling. 
In this letter, we propose a novel variational approach for speckle removal, which involves an expressive image prior model - 
FoE and a highly efficient non-convex optimization algorithm - iPiano, recently proposed in \cite{iPiano}. 
Our proposed method obtains strongly competitive despeckling 
performance w.r.t. the state-of-the-art method - SAR-BM3D \cite{ParrilliPAV12}, 
meanwhile, preserve the property of computational efficiency. 
%As a result, our proposed method leads to a
%variational approaches explicitly incorporate different image prior.
%\vspace*{-0.25cm}
\section{Proposed variational models for despeckling}
In this section, we propose FoE prior based variational models for speckle noise removal and in particular for SAR images. 
We introduce an efficient algorithm to solve the 
corresponding optimization problems. 

Given an observation image $f$ corrupted by multiplicative noise, the FoE prior based despeckling model is defined as 
the following energy minimization problem
\vspace*{-0.25cm}
\begin{equation}\label{general} 
\hat u = \arg\min\limits_{u} E(u, f) = E_{\text{FoE}}(u) + D(u, f) \,,
\vspace*{-0.25cm}
\end{equation}

where $u$ is the underlying unknown image, 
the first term is the FoE prior model, and the second part is the data fidelity term, derived from the 
multiplicative noise model. 
%MAP inference, general formulation
\vspace*{-0.25cm}
\subsection{The FoE prior utilized in our despeckling model}
\begin{figure}[t!]
\begin{center}
    {\includegraphics[width=0.5\textwidth]{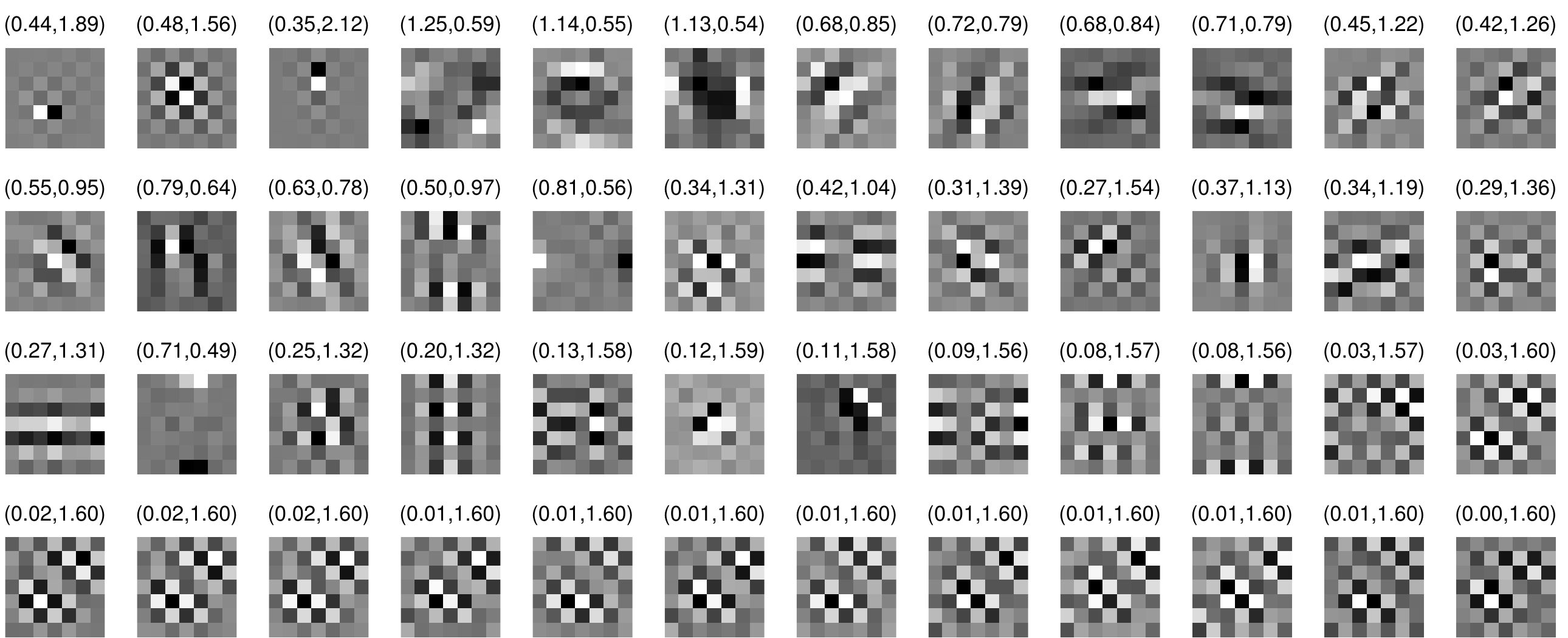}}
\end{center}
\vspace*{-0.3cm}
    \caption{48 learned filters of size $7 \times 7$ exploited in our despeckling model. 
The first number in the bracket is the norm of the filter and the second one is the weight $\theta_i$.}
\label{filters}
\vspace*{-0.5cm}
\end{figure}

The FoE model is defined by a set of linear filters. According to \cite{ChenRP14, ChenPRB13}, 
the student-t distribution based FoE image prior model for an image $u$ is formulated as 
\begin{equation}\label{FOEmodel} 
E_{\text{FoE}}(u) = 
\suml{i=1}{N_f}{\theta_i} \rho(k_i * u) \,, 
\end{equation}
where $\rho(k_i * u) = \suml{p=1}{N} \rho((k_i * u)_p)$, $N$ is the number of pixels in image $u$, 
$N_f$ is the number of linear filters, 
$k_i$ is a set of learned filters with the corresponding weights $\theta_i > 0$, 
$k_i * u$ denotes the convolution of the filter $k_i$ with a two-dimensional image $u$, 
and $\rho(\cdot)$ denotes the non-convex Lorentzian potential function, 
\[
\rho(x) = \text{log}(1 + x^2) \,,
\] 
which is derived from the student-t distribution. Note that the FoE energy is non-convex w.r.t. $u$. 
In this work, we make use of the learned filters of a previous work \cite{ChenRP14}, as shown in Figure~\ref{filters}.

%==========================================================
\vspace*{-0.5cm}
\subsection{Modeling of the data term}
Let $f$ be the observed SAR image amplitude, which follows a Nakagami distribution depending on the underlying 
true image amplitude $u$, the square root of the reflectivity \cite{goodman1975statistical}
\vspace*{-0.1cm}
\[
p(f \vert u) = \frac {2L^L}{\Gamma(L)u^{2L}}f^{2L-1}\text{exp}\left(-\frac {Lf^2}{u^2}\right) \,,
\]
\vspace*{-0.1cm}
with $L$ the number of looks of the image (i.e., number of independent values averaged) and $\Gamma$ 
is the classical Gamma function. 

According to the Gibbs function, this likelihood leads to the following energy term via $E = -\text{log}p(f \vert u)$
\begin{equation}\label{dataterm1} 
D(u, f) = L\cdot (2\text{log} u + \frac {f^2}{u^2} ) \,.
\end{equation}
Combining this data term with the FoE prior model \eqref{FOEmodel}, we reach the following variational model 
\begin{equation}\label{originalNonconvex} 
\arg\min\limits_{u>0} \suml{i=1}{N_f}{\theta_i} \rho(k_i * u) +  \frac \lambda 2 \scal {2\text{log}u+\frac {f^2}{u^2}}{1}\,,
\end{equation} 
where $\scal {\cdot}{\cdot}$ denotes the common inner product. 
Note that the data term is not convex w.r.t. $u > 0$, 
which will generally make the corresponding optimization problem harder to solve. 

There exists an alternative method to define a convex data term by using 
the classical Csisz{\'a}r I-divergence model \cite{csiszar1991least}. 
Although the I-divergence data fitting term typically used in the context of Poisson noise, 
this seemingly inappropriate data term performs very well in the 
TV and TGV regularized variational models for despeckling \cite{SteidlT10, wensen14}. 
Therefore, we also study this variant for data term modeling in this work. 
Following previous works of modeling the SAR image intensity \cite{SteidlT10, wensen14}, 
the I-divergence based data term for the amplitude model is given by
\begin{equation}\label{dataterm2} 
D(u, f) = \frac \lambda 2 (u^2 - 2f^2\text{log}u) \,,
\end{equation}
which is strictly convex w.r.t. $u$ for $u > 0$. Then 
the variational model involving this convex data term is given by 
\begin{equation}\label{variant} 
\arg\min\limits_{u>0} \suml{i=1}{N_f}{\theta_i} \rho(k_i * u) +  \frac \lambda 2 \scal {u^2 - 2f^2\text{log}u}{1}\,. 
\end{equation}
%==========================================================
\vspace*{-1cm}
\subsection{Solving the variational despeckling models}
Due to the non-convexity of the prior term, the proposed variational models impose generally very hard optimization 
problems, especially for the model \eqref{originalNonconvex} with a non-convex data term. 
In this work, we resort to a recently published non-convex optimization algorithm - iPiano \cite{iPiano} to solve them.

The iPiano algorithm is designed for a structured non-smooth non-convex optimization problem, which is composed of 
a smooth (possibly non-convex) function $F$ and a convex (possibly non-smooth) function $G$:
\begin{equation}\label{fplusg} 
\arg\min\limits_{u}H(u) = F(u) + G(u) \,.
\end{equation}
iPiano is an inertial force enhanced forward-backward splitting algorithm with the following basic update rule
\[
u^{n+1} = \left( I + \alpha \partial G \right)^{-1}(u^n - \alpha \nabla F(u^n) + \beta (u^n - u^{n-1})) \,,
\] 
where $\alpha$ and $\beta$ are the step size parameters. 

For the model \eqref{originalNonconvex} with a non-convex data term, 
we can convert the data term to a convex function via commonly used 
logarithmic transformation (i.e., $w = \text{log}u$). Therefore, the minimization problem is rewritten as 
\begin{equation}\label{original} 
\arg\min\limits_{w} \suml{i=1}{N_f}{\theta_i} \rho(k_i * e^w) +  \frac \lambda 2 \scal {2w+f^2e^{-2w}}{1}\,, 
\end{equation}
with $u = e^{w}$. Casting \eqref{original} in the form of \eqref{fplusg}, we see that 
$F(w) = \suml{i=1}{N_f}{\theta_i} \rho(k_i * e^w)$ and $G(w) =\frac \lambda 2 \scal {2w+f^2e^{-2w}}{1}$. 
In order to use the iPiano algorithm, we need to calculate the gradient of $F$ and 
the proximal map w.r.t. $G$. It is easy to check that 
\[
\nabla_w F = \suml{i=1}{N_f}{\theta_i} W K_i^\top \rho'(K_i e^w) \,, 
\]
where $K_i$ is an $N \times N$ highly sparse matrix, implemented as 2D convolution of the image $u$ with filter kernel $k_i$, 
i.e.,, $K_i u \Leftrightarrow k_i * u$, $\rho'(K_i u) = (\rho'((K_i u)_1),\cdots,\rho'((K_i u)_N))^\top \in \R^{N}$, with 
$\rho'(x) = 2x/(1+x^2)$, and $W = \text{diag}(e^w)$. 

The proximal map w.r.t. $G$ is given as the following minimization problem
\begin{equation}\label{subproblem} 
\left( I + \tau \partial G \right)^{-1}(\hat w ) = \arg\min\limits_{w} \frac {\|w - \hat w\|^2_2}{2} + 
\frac {\tau\lambda}{2} \scal {2w+f^2e^{-2w}}{1}\,.
\end{equation}
Instead of using a direct solver for \eqref{subproblem} in terms of the Lambert W function \cite{CorlessGHJK96}, 
we utilized Newton's method. 
We found that this scheme has a quite fast convergence (less than 10 iterations). 

For the variational model \eqref{variant}, 
$F(u) = \suml{i=1}{N_f}{\theta_i} \rho(k_i * u)$, $G(u) =\frac \lambda 2 \scal {u^2 - 2f^2\text{log}u}{1}$. 
Then we have 
\[
\nabla_u F = \suml{i=1}{N_f}{\theta_i} K_i^\top \rho'(K_i u) \,, 
\]
and the proximal map w.r.t. $G$ is given by the following point-wise calculation
\[
\left( I + \tau \partial G \right)^{-1}(\hat u ) \Leftrightarrow 
u_p = \frac{\hat u_p + \sqrt{\hat u_p^2 + 4(1+\tau\lambda)\tau\lambda f_p^2}}{2(1+\tau\lambda)}
\]

\section{Experimental results}
In this section, we evaluated the performance of our proposed variational models based on images corrupted by 
synthetic speckle noise and real noisy SAR images. For synthetic experiments, we calculated the common 
measurements PSNR and SSIM index \cite{WangBSS04}, and compared the results with the state-of-the-art algorithm - 
SAR-BM3D \cite{ParrilliPAV12}. 

\begin{figure}[t!]
\vspace*{-0.175cm}
  \begin{center}
    \subfigure[Noisy image (21.23/0.4267)]{\includegraphics[width=0.24\textwidth]{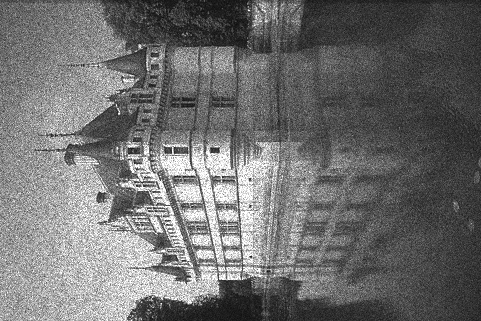}}\hfill
    \subfigure[Model \eqref{original} (29.30/0.8567)]{\includegraphics[width=0.24\textwidth]{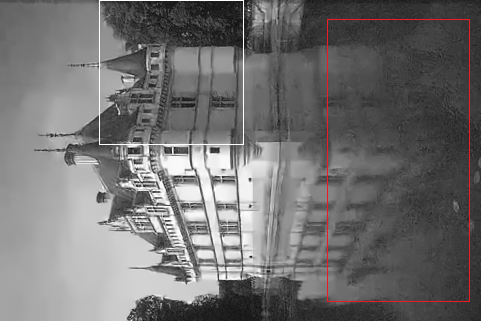}}\\
\vspace*{-0.25cm}
    \subfigure[Model \eqref{variant} (28.40/0.8264)]{\includegraphics[width=0.24\textwidth]{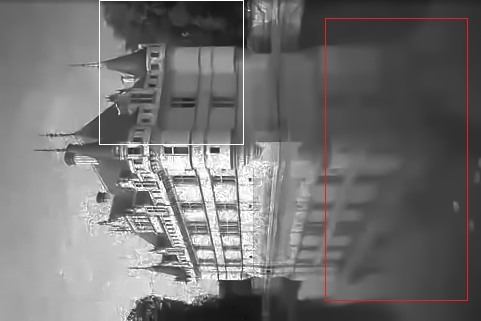}}\hfill
    \subfigure[Combined \eqref{newmodel}(29.62/0.8794)]
{\includegraphics[width=0.24\textwidth]{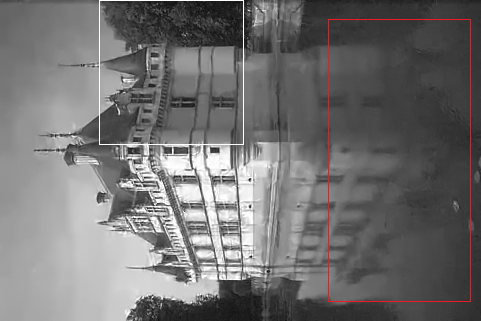}}
    \caption{Despeckling results for a widely used natural image corrupted by multiplicative noise with $L = 8$, using 
our proposed variational models with different data terms. The recovery quality is measured by PSNR/SSIM index.}
\label{preliminary}
  \end{center}
\vspace*{-0.4cm}
\end{figure}
\vspace*{-0.4cm}
\subsection{The influence of data term}
We started with conducting a preliminary despeckling test on a commonly used natural image 
contaminated by multiplicative noise with $L = 8$, using our proposed variational models \eqref{original} and \eqref{variant}. 
We carefully tuned the parameter $\lambda$ to insure that the 
corresponding variational models achieve the best performance. 
The despeckling results are shown in Figure~\ref{preliminary}(b) and (c). 

A first impression from this result is: the variational model with an accurate data fitting term, which is exactly derived from 
the multiplicative noise model can lead to better result than the model with an ``inappropriate'' data term. 
But after having a closer look at the despeckling images, we found an interesting phenomena: these two 
methods possess complementary strengths and failure modes. For example, for the highly textured region (highlighted by 
the white rectangle), \eqref{original} works much better than \eqref{variant}; however, for the homogeneous region 
(highlighted by the red rectangle), \eqref{variant} generates preferable result. Then an intuitive idea arises to integrate 
these two data terms so as to leverage their advantages. 

As a result, a new variational model incorporating these two data terms comes out as follows
\begin{equation}\label{newmodel} 
\arg\min\limits_{w} E_{\text{FoE}}(e^w) +  \frac {\lambda_1} {2} (2w+f^2e^{-2w}) + 
\frac {\lambda_2} {2} (e^{2w} - 2f^2w)\,, 
\end{equation}
with $u = e^{w}$. The data fitting term is still convex, and we can utilize iPiano to solve it. The proximal map w.r.t. $G$ 
is also solved using Newton's method. 

We manually tuned the parameters $\lambda_1$ and $\lambda_2$, and conducted the same despeckling 
experiment. The final result is shown in Figure~\ref{preliminary}(d). One can see that the combined model 
indeed leads to significant improvement in terms of both PSNR and SSIM index value. From now on, we will use the 
combined model \eqref{newmodel} for despeckling experiments. 
%$\lambda_1 = 550$ and $\lambda_2 = 0.02$,
%=====================================================================
\vspace*{-0.4cm}
\subsection{Results on synthetic speckle noise}
\begin{figure}[t!]
  \begin{center}
    {\includegraphics[width=0.16\textwidth]{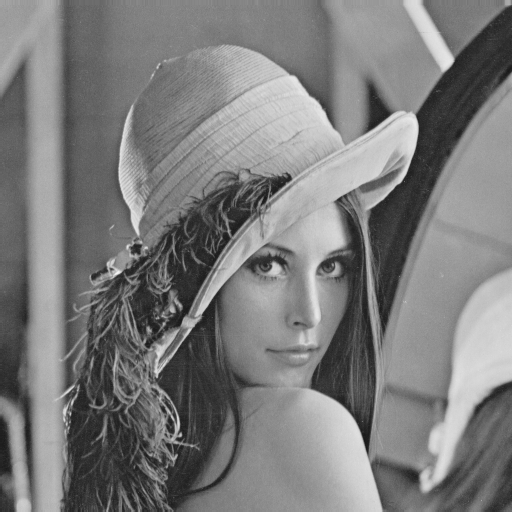}}\hfill
    {\includegraphics[width=0.16\textwidth]{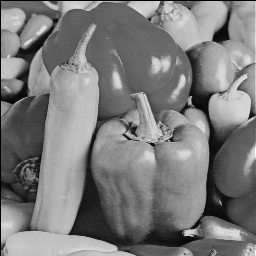}}\hfill
    {\includegraphics[width=0.16\textwidth]{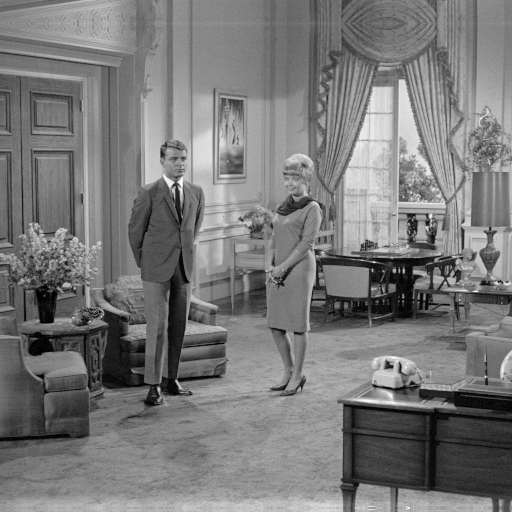}}
    \caption{Standard test images of size $256 \times 256$ (Couple, Lenna and Peppers).}
\label{testimages}
  \end{center}
\vspace*{-0.25cm}
\end{figure}
We first evaluated the performance of our proposed variational model \eqref{newmodel} 
for despeckling based on synthetic noisy images. 
The results are compared to the best published despeckling algorithm - SAR-BM3D \cite{ParrilliPAV12}. 
First of all, we considered three standard test images widely used in image processing community, see in Figure~\ref{testimages}. 
The despeckling results at three representative numbers of look $L = 1, 3, 8$ are summarized in Table~\ref{results}. 
We made the following empirical choice for the parameters $\lambda_1$ and $\lambda_2$ for different $L$: 
if $L = 8$, $\lambda_1 = 550, \lambda_2 = 0.02$; if $L = 3$, $\lambda_1 = 310, \lambda_2 = 0.008$; and if 
$L = 1$, $\lambda_1 = 160, \lambda_2 = 0.004$. A practical guideline to tune $\lambda_1$ and $\lambda_2$, as well as 
the implementation of the proposed approach can be found at our homepage \cite{gpu4vision}. 

\begin{table}[t!]
\centering
\begin{tabular}{|c |c |c ||c |c ||c |c |}
%\Xhline{1.2pt}
\cline{1-7}
&\multicolumn{2}{c|}{Lenna} &\multicolumn{2}{c|}{Peppers} &\multicolumn{2}{c|}{Couple}\\
\cline{1-7}
\multirow{2}*{$L = 8$}
&31.29 & 0.8948 &29.69 &0.8693 &29.01 &0.8416\\
\cline{2-7}
&\color{blue}31.38 & \color{blue}0.8968 &\color{blue}30.64 &\color{blue}0.8803 &\color{blue}29.23 &\color{blue}0.8368\\
\hline
\hline
\multirow{2}*{$L = 3$}
&28.81 & 0.8453 &27.48 &0.8233 &26.60 &0.7559\\
\cline{2-7}
&\color{blue}28.85 & \color{blue}0.8541 &\color{blue}28.38 &\color{blue}0.8409 &\color{blue}26.52 &\color{blue}0.7510\\
\hline
\hline
\multirow{2}*{$L = 1$}
&25.92 & 0.7432 &24.95 &0.7495 &23.98 &0.6210\\
\cline{2-7}
&\color{blue}25.85 & \color{blue}0.7771 &\color{blue}25.34 &\color{blue}0.7770 &\color{blue}23.79&\color{blue}0.6023\\
\cline{1-7}
\end{tabular}
\vspace*{0.1cm}
\caption{despeckling results of SAR-BM3D \cite{ParrilliPAV12} and our approach. {\color{blue}{Our}} 
results are marked with {\color{blue}{blue}} color (with{\color{blue}{ \eqref{newmodel}}}). 
The results are reported with PSNR and SSIM values.}
\label{results}
\vspace*{-0.75cm}
\end{table}

\begin{figure}[t!]
\vspace*{-0.25cm}
  \begin{center}
    \subfigure[$L$ = 1 (12.10/0.1595)]{\includegraphics[width=0.16\textwidth]{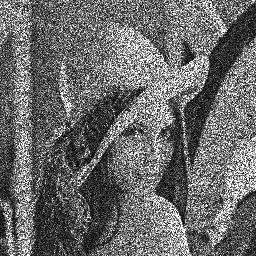}}\hfill
    \subfigure[{\cite{ParrilliPAV12}(25.92/0.7432)}]{\includegraphics[width=0.16\textwidth]{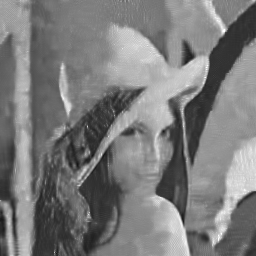}}\hfill
    \subfigure[Ours (25.85/0.7771)]{\includegraphics[width=0.16\textwidth]{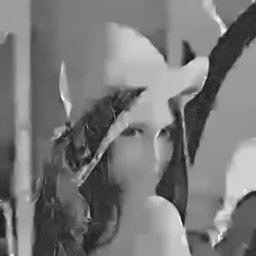}}\\
\vspace*{-0.25cm}
    \subfigure[$L$ = 3 (16.48/0.2939)]{\includegraphics[width=0.16\textwidth]{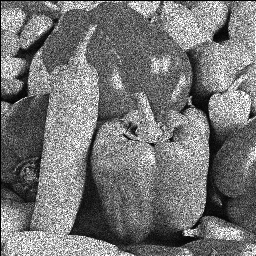}}\hfill
    \subfigure[{\cite{ParrilliPAV12}(27.48/0.8233)}]{\includegraphics[width=0.16\textwidth]{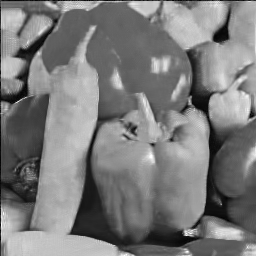}}\hfill
    \subfigure[Ours (28.38/0.8409)]{\includegraphics[width=0.16\textwidth]{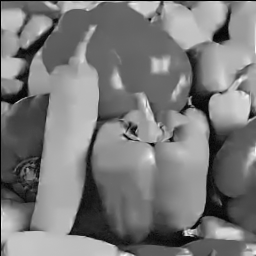}}\\
\vspace*{-0.25cm}
    \subfigure[$L$ = 8 (23.19/0.6981)]{\includegraphics[width=0.16\textwidth]{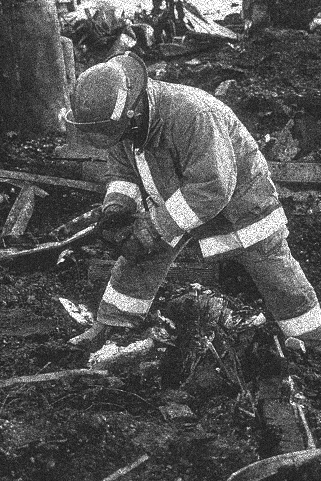}}\hfill
    \subfigure[{\cite{ParrilliPAV12}(28.06/0.8663)}]{\includegraphics[width=0.16\textwidth]{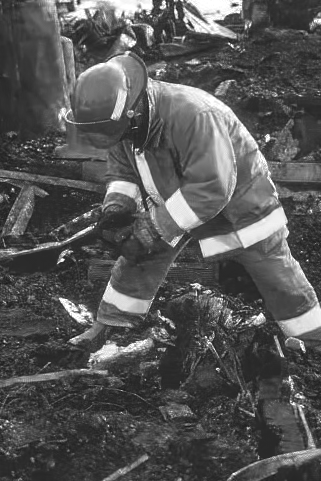}}\hfill
    \subfigure[Ours (28.39/0.8647)]{\includegraphics[width=0.16\textwidth]{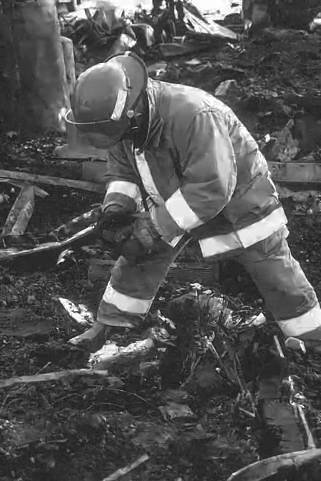}}\\
\vspace*{-0.1cm}
    \caption{Performance comparison to state-of-the-art algorithm - SAR-BM3D \cite{ParrilliPAV12} for different $L$. 
The results are reported by PSNR/SSIM index.}
\label{despeckling}
  \end{center}
\vspace*{-0.5cm}
\end{figure}

From Table~\ref{results}, one can see that these two competing approaches generate very similar results, with 
almost identical PSNR and SSIM values. 
We present three despeckling examples obtained by these two algorithms in Figure~\ref{despeckling}. 

As we know, the despeckling performance of one particular method varies greatly for different image contents, in order to 
make a comprehensive comparison, we conducted additional despeckling experiments over a standard test dataset - 68 
Berkeley test images identified by Roth and Black \cite{RothFOE2009}, which is widely used for AWGN denoising test. 
All the results were computed per image and then averaged over the test dataset. 
We present the results in Table~\ref{68testimages}. 
Again, two competing algorithms behave similarly, which implies that our proposed 
variational model based on the FoE prior is on par with the best published despeckling algorithm - SAR-BM3D. 

In experiments, we found that our method introduces block-type artifacts for the case of low $L$, \eg, 
$L =1$ in Figure~\ref{despeckling}(c). The main reason is that our method is a local model, which becomes less effective to infer 
the underlying structure solely from the local neighborhoods, if the input image is too noisy. In this case, the advantage of non-locality 
comes through.

\begin{table}[t!]
\vspace*{0.175cm}
\centering
\begin{tabular}{|c |c |c ||c |c ||c |c |}
\cline{1-7}
&\multicolumn{2}{c|}{$L = 8$} &\multicolumn{2}{c|}{$L = 3$} &\multicolumn{2}{c|}{$L = 1$}\\
\cline{1-7}
Noisy &21.61 & 0.5355 &17.42 &0.3778 &12.95 &0.2231\\
\hline
\hline
SAR-BM3D &29.35 & 0.8520  &27.12 &0.7756 &24.85 &0.6757\\
\hline
\hline
Ours &29.54 & 0.8481 &27.07 &0.7628 &24.65 &0.6691 \\
\hline
\end{tabular}
\vspace*{0.1cm}
\caption{despeckling results on 68 Berkeley test images. 
The results are reported with average PSNR and SSIM values.}
\label{68testimages}
\vspace*{-0.5cm}
\end{table}
%====================================================

\vspace*{-0.1cm}
\subsection{Results on a real SAR image}
\begin{figure}[t!]
\vspace*{-0.175cm}
  \begin{center}
    \subfigure[Noisy image, $L = 5$]{\includegraphics[width=0.16\textwidth]{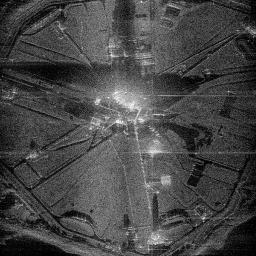}}\hfill
    \subfigure[SAR-BM3D\cite{ParrilliPAV12}]{\includegraphics[width=0.16\textwidth]{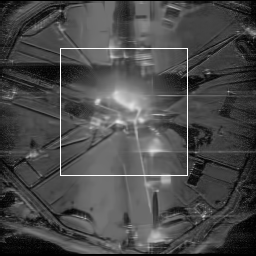}}\hfill
    \subfigure[Ours]{\includegraphics[width=0.16\textwidth]{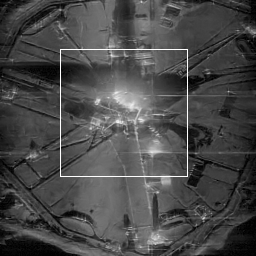}}\\
\vspace*{-0.25cm}
    \subfigure[PPBit\cite{DeledalleDT09}]{\includegraphics[width=0.16\textwidth]{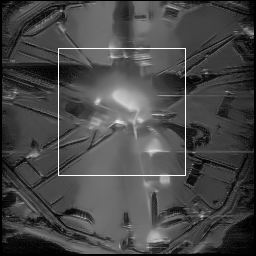}}
    \subfigure[TGV based\cite{wensen14}]{\includegraphics[width=0.16\textwidth]{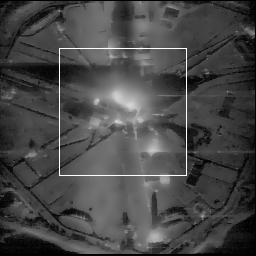}}
    \caption{Performance comparison of different algorithms on a real SAR image.}
\label{realSAR}
  \end{center}
\vspace*{-0.5cm}
\end{figure}
In order to demonstrate the effectiveness of our proposed method on real SAR image despeckling task, we 
conducted a despeckling test on a real noisy SAR image, 
which is taken by the radar system equipped with the JSTARS aircraft \cite{SAR} (the number of looks $L = 5$). 
%exploited in a recent work \cite{wensen14}. 
For this experiment, we set $\lambda_1 = 50, \lambda_2 = 0.15$ for our model. 

The results of different algorithms are shown in Figure~\ref{realSAR}. 
One can see that the tiny structures in the image, especially the region highlighted by the white rectangle, are 
well-preserved by our approach after despeckling; however, they are smoothed out to different extents by 
other algorithms. 
\vspace*{-0.25cm}
\subsection{Run time}
Efficiency is also an important aspect for real SAR despeckling task. 
On the server platform of Intel X5675 3.07GHz, 
for images of size $481 \times 321$ exploited in our experiments, a typical run time of the SAR-BM3D algorithm is about 
65.4s, which can be reduced to 5.6s, by its improved version - FANS \cite{CozzolinoPSPV14}, at the expense of 
slight performance degradation. Our method typically consumes 27s with a pure Matlab code. 

However, our model is a local model, which solely contains convolution of linear filters with an image, 
in contrast to non-local models, \eg, SAR-BM3D and its variant FANS. Therefore, our model is 
well-suited to GPU parallel computation. 
Our GPU implementation based on a NVIDIA Geforce GTX 680 
accelerates the inference procedure significantly; for a despeckling task with $L = 8$, it typically takes 0.6s for images 
of size $512 \times 512$, 0.41s for $481 \times 321$ and 0.2s for $256 \times 256$.

\section{Conclusion}
In this letter, we have proposed a novel variational model for speckle noise reduction, 
based on an expressive image prior model - FoE model. Our new variational model poses a generally 
demanding non-convex optimization problem and we have used the recently proposed algorithm - 
iPiano to solve it efficiently. Numerical results on synthetic images corrupted by speckle noise and a real SAR image 
demonstrate that the performance of our method is clearly on par with the best published despeckling algorithm - 
SAR-BM3D. Furthermore, our model comes along with the additional advantage of simplicity and therefore well-suited to 
GPU programming. The GPU based implementation can conduct despeckling in less than 1s with state-of-the-art 
performance. 

%Future work includes training the specialized FoE prior model for SAR images based on images of scenes 
%resembling real SAR images, \eg, urban scenes of large scale. 
\bibliographystyle{ieee}
\bibliography{despeckle}

\begin{thebibliography}{10}\itemsep=-1pt

\bibitem{SAR}
\url{http://academic.emporia.edu/aberjame/student/graves1/project.html}.

\bibitem{gpu4vision}
\url{http://gpu4vision.icg.tugraz.at/}.

\bibitem{ArgentiBLA12}
F.~Argenti, T.~Bianchi, A.~Lapini, and L.~Alparone.
\newblock Fast {MAP} despeckling based on laplacian-gaussian modeling of
  wavelet coefficients.
\newblock {\em IEEE Geosci. Remote Sensing Lett.}, 9(1):13--17, 2012.

\bibitem{AubertA08}
G.~Aubert and J.-F. Aujol.
\newblock A variational approach to removing multiplicative noise.
\newblock {\em SIAM Journal of Applied Mathematics}, 68(4):925--946, 2008.

\bibitem{BianchiAA08}
T.~Bianchi, F.~Argenti, and L.~Alparone.
\newblock Segmentation-based {MAP} despeckling of {SAR} images in the
  undecimated wavelet domain.
\newblock {\em IEEE T. Geoscience and Remote Sensing}, 46(9):2728--2742, 2008.

\bibitem{BrediesKP10}
K.~Bredies, K.~Kunisch, and T.~Pock.
\newblock Total generalized variation.
\newblock {\em SIAM J. Imaging Sciences}, 3(3):492--526, 2010.

\bibitem{ChenPRB13}
Y.~Chen, T.~Pock, R.~Ranftl, and H.~Bischof.
\newblock Revisiting loss-specific training of filter-based mrfs for image
  restoration.
\newblock In {\em GCPR}, pages 271--281, 2013.

\bibitem{ChenRP14}
Y.~Chen, R.~Ranftl, and T.~Pock.
\newblock Insights into analysis operator learning: From patch-based sparse
  models to higher order {MRFs}.
\newblock {\em IEEE Transactions on Image Processing}, 23(3):1060--1072, 2014.

\bibitem{CorlessGHJK96}
R.~M. Corless, G.~H. Gonnet, D.~E.~G. Hare, D.~J. Jeffrey, and D.~E. Knuth.
\newblock On the lambert {W} function.
\newblock {\em Adv. Comput. Math.}, 5(1):329--359, 1996.

\bibitem{CozzolinoPSPV14}
D.~Cozzolino, S.~Parrilli, G.~Scarpa, G.~Poggi, and L.~Verdoliva.
\newblock Fast adaptive nonlocal {SAR} despeckling.
\newblock {\em IEEE Geosci. Remote Sensing Lett.}, 11(2):524--528, 2014.

\bibitem{csiszar1991least}
I.~Csisz{\'a}r.
\newblock Why least squares and maximum entropy? an axiomatic approach to
  inference for linear inverse problems.
\newblock {\em The annals of statistics}, 19(4):2032--2066, 1991.

\bibitem{BM3D}
K.~Dabov, A.~Foi, V.~Katkovnik, and K.~O. Egiazarian.
\newblock Image denoising by sparse 3-d transform-domain collaborative
  filtering.
\newblock {\em IEEE Transactions on Image Processing}, 16(8):2080--2095, 2007.

\bibitem{DeledalleDT09}
C.-A. Deledalle, L.~Denis, and F.~Tupin.
\newblock Iterative weighted maximum likelihood denoising with probabilistic
  patch-based weights.
\newblock {\em IEEE Transactions on Image Processing}, 18(12):2661--2672, 2009.

\bibitem{wensen14}
W.~Feng, H.~Lei, and Y.~Gao.
\newblock Speckle reduction via higher order total variation approach.
\newblock {\em Image Processing, IEEE Transactions on}, 23(4):1831--1843, April
  2014.

\bibitem{goodman1975statistical}
J.~W. Goodman.
\newblock Statistical properties of laser speckle patterns.
\newblock In {\em Laser speckle and related phenomena}, pages 9--75. Springer,
  1975.

\bibitem{kuan1987adaptive}
D.~T. Kuan, A.~Sawchuk, T.~C. Strand, and P.~Chavel.
\newblock Adaptive restoration of images with speckle.
\newblock {\em Acoustics, Speech and Signal Processing, IEEE Transactions on},
  35(3):373--383, 1987.

\bibitem{lee1980digital}
J.-S. Lee.
\newblock Digital image enhancement and noise filtering by use of local
  statistics.
\newblock {\em Pattern Analysis and Machine Intelligence, IEEE Transactions
  on}, (2):165--168, 1980.

\bibitem{LiWHY13}
G.-T. Li, C.-L. Wang, P.-P. Huang, and W.-D. Yu.
\newblock {SAR} image despeckling using a space-domain filter with alterable
  window.
\newblock {\em IEEE Geosci. Remote Sensing Lett.}, 10(2):263--267, 2013.

\bibitem{iPiano}
P.~Ochs, Y.~Chen, T.~Brox, and T.~Pock.
\newblock i{P}iano: Inertial {P}roximal {A}lgorithm for {N}onconvex
  {O}ptimization.
\newblock {\em SIAM Journal on Imaging Sciences}, 7(2):1388--1419, 2014.

\bibitem{ParrilliPAV12}
S.~Parrilli, M.~Poderico, C.~V. Angelino, and L.~Verdoliva.
\newblock A nonlocal {SAR} image denoising algorithm based on {LLMMSE} wavelet
  shrinkage.
\newblock {\em IEEE T. Geoscience and Remote Sensing}, 50(2):606--616, 2012.

\bibitem{RothFOE2009}
S.~Roth and M.~J. Black.
\newblock Fields of experts.
\newblock {\em International Journal of Computer Vision}, 82(2):205--229, 2009.

\bibitem{SteidlT10}
G.~Steidl and T.~Teuber.
\newblock Removing multiplicative noise by douglas-rachford splitting methods.
\newblock {\em Journal of Mathematical Imaging and Vision}, 36(2):168--184,
  2010.

\bibitem{WangBSS04}
Z.~Wang, A.~C. Bovik, H.~R. Sheikh, and E.~P. Simoncelli.
\newblock Image quality assessment: from error visibility to structural
  similarity.
\newblock {\em IEEE Transactions on Image Processing}, 13(4):600--612, 2004.

\bibitem{YunW12}
S.~Yun and H.~Woo.
\newblock A new multiplicative denoising variational model based on $m$th root
  transformation.
\newblock {\em IEEE Transactions on Image Processing}, 21(5):2523--2533, 2012.

\end{thebibliography}

\end{document}